\documentclass{llncs}
\usepackage{amssymb}

\title{Vicious Circle Principle and Formation of Sets in ASP Based Languages}

  \author{Michael Gelfond and Yuanlin Zhang}

  \institute{Texas Tech University, Lubbock, TX, USA \\
	  \email{\{michael.gelfond, y.zhang\}@ttu.edu}
  }

\newcommand{\hide}[1]{}
\newcommand{\otherquestions}[1]{}

\def\st{\medskip\noindent}

\def\st{\medskip\noindent}

\begin{document}

\maketitle
\begin{abstract}
The paper continues the investigation of Poincare and Russel's Vicious Circle Principle
(VCP) in the context of the design of logic programming
languages with sets.  We expand previously introduced language
$\mathcal{A}log$ with aggregates by allowing infinite sets and
several additional set related constructs useful for knowledge
representation and teaching. In addition, we propose an
alternative formalization of the original VCP and incorporate it into the
semantics of new language, $\mathcal{S}log^+$, which allows more liberal
construction of sets and their use in programming rules. We show that,
for programs without disjunction and infinite sets, the formal semantics of aggregates in $\mathcal{S}log^+$ coincides with 
that of several other known languages. Their intuitive and formal semantics,
however, are based on quite different ideas and seem to be more involved than that of $\mathcal{S}log^+$. 
 
\end{abstract}
\section{Introduction}
This paper is the continuation of work started in \cite{GelfondZ14} with introduction 
of $\mathcal{A}log$ -- a version of Answer Set Prolog (ASP) with
aggregates. The semantics of $\mathcal{A}log$ combines the Rationality
Principle of ASP \cite{GelK14} with the adaptation of the Vicious Circle Principle (VCP)
introduced by Poincare and Russel \cite{poin1906,Russell} in their attempt to resolve
paradoxes of set theory.
In $\mathcal{A}log$, the latter is used to deal with formation of sets
and their legitimate use in program rules. 
To understand the difficulty addressed by $\mathcal{A}log$ consider
the following programs:

\begin{example}\label{e1} 
$P_0$ consisting of a rule:
\begin{verbatim}
p(1) :- card{X: p(X)} != 1. 
\end{verbatim}

\noindent
$P_1$ consisting of rules:
\begin{verbatim}
p(1) :- p(0). 
p(0) :- p(1). 
p(1) :- card{X: p(X)} != 1. 
\end{verbatim}

\noindent
$P_2$ consisting of rules:

\begin{verbatim}
p(1) :- card{X: p(X)} >= 0. 
\end{verbatim}
Even for these seemingly simple programs, there are different opinions 
about their meaning. To the best of our knowledge all ASP based
semantics, including that of  \cite{FaberPL11,SonP07,GelfondZ14})
view $P_0$ as a bad specification. It is inconsistent, i.e., has no
answer sets. Opinions differ, however, about the meaning of the other
two programs. \cite{FaberPL11} views
$P_1$ as a reasonable specification having one answer 
set -- $\{p(0),p(1) \}$. According to \cite{SonP07,GelfondZ14}
$P_1$ is inconsistent.
According to most semantics $P_2$ has one answer set, $\{p(1)\}$. 
$\mathcal{A}log$, however, views it as inconsistent. 
\end{example}

\noindent
As in the naive set theory, the difficulty in interpretations seems to
be caused by self-reference. 
In both $P_1$ and $P_2$, the definition of $p(1)$
references the set described in terms of $p$.
It is, of course, not entirely clear how this type of differences can
be resolved. Sometimes, further analysis can find convincing arguments
in favor of one of the proposals. Sometimes, the analysis discovers
that different approaches really model different language or world
phenomena and are, hence, all useful in different contexts. 
We believe that the difficulty can be greatly alleviated 
if the designers of
the language provide its users with as clear intuitive meaning
of the new constructs as possible.
Accordingly,  the \emph{set name} construct $\{X:p(X)\}$ of
$\mathcal{A}log$
denotes \emph{the set of all objects believed by the rational 
  agent associated with the program to satisfy property $p$}.
(This reading is in line with the epistemic view of ASP connectives
shared by the authors.)
The difficulties with self-reference in $\mathcal{A}log$ are resolved  
by putting the following intuitive restriction on the formation of
sets\footnote{It is again similar to set theory where the difficulty
  is normally avoided by restricting comprehension axioms guaranteeing
  existence of sets denoted by expressions of the form
  $\{X:p(X)\}$. In ASP such restrictions are
  encoded in the definition of answer sets.}:

\st
\emph{An expression $\{X:p(X)\}$ denotes a set $S$ only if for every $t$ rational belief in 
$p(t)$ can be established without a reference to $S$ }, or
equivalently, 
\emph{the reasoner's belief in $p(t)$ can not depend on existence of a
  set denoted by $\{X:p(X)\}$}.

\st
We view this restriction as a possible interpretation of VCP and refer
to it as \emph{Strong  VCP}.
Let us illustrate the intuition behind $\mathcal{A}log$ set constructs.
\begin{example}\label{e2}
{\rm Let us consider programs from Example \ref{e1}.  
$P_0$ clearly has no answer set since $\emptyset$ does not satisfy
its rule and there is no justification for believing in $p(1)$.
$P_1$ is also inconsistent. To see that notice that 
the first two rules of the program limit our  
possibilities to $A_1 = \emptyset$ and $A_2=\{p(0), p(1)\}$. In the first  
case $\{X:p(X)\}$ denotes $\emptyset$. But this  
contradicts the last rule of the program. $A_1$ cannot be an answer  
set of $P_1$. In $A_2$,  $\{X:p(X)\}$ denotes
$S=\{0, 1\}$. But this violates our form of VCP since the  
reasoner's beliefs in both, $p(0)$ and $p(1)$, cannot be established  
without  reference to $S$. $A_2$ is not an answer set either.  
Now consider program $P_2$. There are two candidate answer
sets\footnote{By a candidate answer set we mean a consistent set of
  ground regular literals satisfying the rules of the program.}: $A_1 =
\emptyset$ and $A_2 = \{p(1)\}$. In $A_1$, $S = \emptyset$ which contradicts the rule.  
In $A_2$, $S=\{1\}$ but this
would contradict the $\mathcal{A}log$'s VCP.  
The program is inconsistent\footnote{There is a common argument for the semantics in which $\{p(1)\}$ would be
the answer set of $P_2$: ``Since $card\{X: p(X)\} \geq 0$ is always true
it can be dropped from the rule without changing the rule's
meaning''. But the argument assumes existence of the set denoted by $\{X:p(X)\}$ which is
not always the case in $\mathcal{A}log$.}.
}
\end{example}
We hope that the examples are sufficient to show how the informal
semantics of $\mathcal{A}log$ 
can give a programmer some guidelines in avoiding formation of sets 
problematic from the standpoint of VCP. 
In what follows we 
\begin{itemize}
\item Expand $\mathcal{A}log$ by allowing infinite sets and
several additional set related constructs useful for knowledge
representation and teaching. 
\item Propose an
alternative formalization of the original VCP and incorporate it into the
semantics of new language, $\mathcal{S}log^+$, which allows more liberal
construction of sets and their use in programming rules. (The name of
the new language is explained by its close relationship with language
$\mathcal{S}log$ \cite{SonP07} -- see Theorem 2).
\item Show that, for programs without disjunction and infinite sets,
  the formal semantics of aggregates in $\mathcal{S}log^+$ coincides with 
that of several other known languages. Their intuitive and formal semantics,
however, are based on quite different ideas and seem to be more involved than that of $\mathcal{S}log^+$.
\item Prove some basic properties of programs in (extended)
  $\mathcal{A}log$ and $\mathcal{S}log^+$. 
\end{itemize}

\section{Syntax and Semantics of $\mathcal{A}log$}
In what follows we retain the name $\mathcal{A}log$ for the new
language and refer to the earlier version as ``original $\mathcal{A}log$''.
\subsection{Syntax}
Let $\Sigma$ be a (possibly sorted) signature with a finite collection 
of predicate and function symbols and (possibly infinite) collection of
object constants, and let $\mathcal{A}$ be a finite 
collection of symbols used to denote functions from
sets of 
terms of $\Sigma$ into integers.
Terms and literals over signature 
$\Sigma$ are defined as usual and referred to as \emph{regular}. 
Regular terms are called \emph{ground} if they contain no variables 
and no occurrences of symbols for arithmetic functions. Similarly for 
literals. 
We refer to an expression
\begin{equation}\label{set-name}
\{\bar{X}:cond\}  
\end{equation}
where $cond$ is a finite collection of regular 
literals and $\bar{X}$ is the list of variables occurring in $cond$,
as a \emph{set name}.
It is read as \emph{the set of all objects of the 
  program believed to satisfy $cond$}.  
Variables from $\bar{X}$ are often referred to as {\em set variables}. 
An occurrence of a set variable in (\ref{set-name}) 
is called \emph{bound} within (\ref{set-name}). 
Since treatment of variables in extended $\mathcal{A}log$ is the same as in the 
original language we limit our attention to programs in which every
occurrence of a variable is bound. Rules containing non-bound
occurrences of variables are considered as shorthands
for their ground instantiations (for details see \cite{GelfondZ14}).

\noindent
A \emph{set atom} of $\mathcal{A}log$ is an expression of the form
\begin{equation}\label{aggr-atom1}
f_1(S_1) \odot f_2(S_2)
\end{equation}
or
\begin{equation}\label{aggr-atom2}
f(S)  \odot k
\end{equation}
where $f$, $f_1,f_2$ are functions from ${\cal A}$, $S$, $S_1$, $S_2$ are set
names, $k$ is a number, and $\odot$ is
an arithmetic relation $>, \geq, <, \leq,=$ or !=, or of the form 
\begin{equation}\label{set-atom}
S_1 \otimes S_2 
\end{equation}
where $\otimes$ is $\subset, \subseteq$, or $=$.
We often write $f(\{\bar{X}:p(\bar{X})\})$ as $f\{\bar{X}:p(\bar{X})\}$
and $\{\bar{X} : p(\bar{X})\} \otimes S$ and $S \otimes \{\bar{X} : p(\bar{X})\}$ 
as $p \otimes S$ and $S \otimes p$ respectively.
Regular and set atoms are referred to as
  \emph{atoms}. 
A \emph{rule} of 
$\mathcal{A}log$ is an expression of the form 
\begin{equation}\label{rule}
head \leftarrow body
\end{equation}
where $head$ is a disjunction of regular literals or a set atom of the
form $p \subseteq S$, $S \subseteq q$, or $p = S$,
and $body$ is a collection of regular literals (possibly preceded by
$not$) and set atoms.
A rule with set atom in the head is called \emph{set introduction rule}.
Note that both head and body of a rule can be infinite. 
All parts of $\mathcal{A}log$ rules, including $head$, can be empty. 
A \emph{program} of $\mathcal{A}log$ is a collection of 
$\mathcal{A}log$'s rules. 
\subsection{Semantics}
To define the semantics of $\mathcal{A}log$ programs we first notice
that the \emph{standard definition of answer set from \cite{gl91} is
applicable to programs with infinite rules}. Hence we already
have the definition of answer set 
for $\mathcal{A}log$ programs not containing
occurrences of set atoms.
We also need the satisfiability relation for set atoms.
Let $A$ be a set of ground regular literals.
If $f(\{\bar{t}:cond(\bar{t}) \subseteq A\})$ is defined then 
$f(\{\bar{X}:cond\}) \geq k$ is satisfied by $A$ (is \emph{true} in $A$)
iff $f(\{\bar{t}:cond(\bar{t}) \subseteq A\}) \geq k$.
Otherwise, $f(\{\bar{X}:cond\}) \geq k$
is falsified (is \emph{false} in $A$).
If $f(\{\bar{t}:cond(\bar{t})
\subseteq A\})$ is not defined then $f(\{\bar{X}:cond\}) \geq k$ is
\emph{undefined} in $A$. 
(For instance, atom $card\{X:p(X)\} \geq 0$ is undefined in $A$ if 
$A$ contains an infinite collection of atoms formed by $p$.) 
Similarly for other set atoms.
Finally a rule is \emph{satisfied} by $S$ if its head is \emph{true}
in $S$ or its body is \emph{false} or
\emph{undefined} in $S$. 
\subsubsection{Answer Sets for Programs without Set Introduction Rules.}
To simplify the presentation we first give the definition of
answer sets for programs whose rules contain no set atoms in their heads.
First we need the following definition:
\begin{definition}[Set Reduct of $\mathcal{A}log$]\label{reduct1}
Let $\Pi$  be a ground program of $\mathcal{A}log$.
The \emph{set reduct} of $\Pi$ with respect to a set of 
ground regular literals $A$ is obtained from $\Pi$ by 
\begin{enumerate}
\item removing rules containing set atoms
  which are \emph{false} or \emph{undefined} in $A$. 
\item replacing every remaining set atom $SA$ by the union of $cond(\bar{t})$ such 
  that $\{\bar{X} : cond(\bar{X})\}$ occurs in $SA$ and $cond(\bar{t})
  \subseteq A$.
\end{enumerate}
\end{definition}
The first clause of the definition removes rules useless because of 
the truth values of their aggregates in $A$. 
The next clause reflects the principle of avoiding 
vicious circles.
Clearly, set reducts do not contain set atoms.
\begin{definition}[Answer Set]\label{ans-set}
A set $A$ of ground regular literals over the signature of a ground
$\mathcal{A}log$ program $\Pi$ is an 
\emph{answer set} of $\Pi$ if $A$ is an answer set of the set
reduct of $\Pi$ with respect to $A$. 
\end{definition}
It is easy to see that for programs of the original $\mathcal{A}log$
our definition coincides with the old one. Next several examples demonstrate the behavior of our semantics for 
programs not covered by the original syntax. 

\st{\bf Infinite Universe}
\begin{example}[Aggregates on infinite sets] \label{new1}
Consider a program $E_1$ consisting of the following rules:
\begin{verbatim}
even(0). 
even(I+2) :- even(I). 
q :- min{X : even(X)} = 0. 
\end{verbatim}
\noindent 
It is easy to see that the program has one answer set,
$S_{E_1} = \{q,even(0),even(2),\dots\}$. Indeed, the reduct of $E_1$ with respect to 
$S_{E_1}$ is the infinite collection of rules 
\begin{verbatim}
even(0). 
even(2) :- even(0). 
... 
q :- even(0),even(2),even(4)... 
\end{verbatim}
\noindent
The last rule has the infinite body constructed in the last step of 
definition \ref{reduct1}. Clearly, $S_{E_1}$ is a subset minimal collection 
of ground literals satisfying the rules of the reduct (i.e. its answer
set). Hence  $S_{E_1}$ is an answer set of $E_1$.
\end{example}
\begin{example}[Programs with undefined aggregates]\label{new2} 
Now consider a program $E_2$ consisting of the rules:
\begin{verbatim} 
even(0).  
even(I+2) :- even(I).  
q :- card{X : even(X)} > 0. 
\end{verbatim}

\noindent
This program has one answer set, $S_{E_2}= \{even(0),even(2),\dots\}$.  
Since our aggregates range over natural
numbers, the aggregate $card$ is not defined on the set $card\{t :
even(t) \in S_{E_2}\}$. This means that the body of the last rule is
undefined. According to clause one of definition \ref{reduct1} this rule is removed.
The reduct of $E_2$ with respect to $S_{E_2}$ is
\begin{verbatim}
even(0). 
even(2) :- even(0). 
even(4) :- even(2). 
... 
\end{verbatim}   
Hence $S_{E_2}$ is the answer set of $E_2$.\footnote{Of course this
  is true only because of our (somewhat arbitrary) decision to limit
  aggregates of $\mathcal{A}log$ to those ranging over natural
  numbers. We could, of course, allow aggregates mapping sets into
  ordinals. In this case the body of the last rule of $E_2$ will be
  defined and the only answer set of $E_2$ will be $S_{E_1}$.}
It is easy to check that, since every set $A$ satisfying the rules of $E_2$ must 
contain all even numbers, $S_{E_1}$ is the only answer set.

\end{example}

\noindent
{\bf Programs with Set Atoms in the Bodies of Rules}
\begin{example}[Set atoms in the rule body]
Consider a knowledge base containing two complete lists of atoms:
\begin{verbatim}
taken(mike,cs1).  taken(mike,cs2).  taken(john,cs2).         
required(cs1).    required(cs2). 
\end{verbatim}
Set atoms allow for a natural definition of the new relation, 
$ready\_to\_graduate(S)$, which holds if student $S$ has taken all the 
required classes from the second list:   

\st {\tt 
ready\_to\_graduate(S) :- \{C: required(C)\} $\subseteq$ \{C:taken(S,C)\}. }

\st The intuitive meaning of the rule is reasonably clear. 
The program consisting of this rule and the closed world assumption:
\begin{verbatim}
-ready_to_graduate(S) :- not ready_to_graduate(S) 
\end{verbatim}
implies that Mike is ready to graduate while John is not. 
If the list of classes taken by a student is incomplete the closed 
world assumption should be removed but the first 
rule still can be useful to determine people who are definitely ready 
to graduate. 
Even though the story can be represented in 
ASP without the set atoms, such representations are substantially less
intuitive and less elaboration tolerant. Here is a simplified example of
alternative representation suggested to the authors by a third
party:

\st {\tt 
ready\_to\_graduate :- not -ready\_to\_graduate. }

\st {\tt 
-ready\_to\_graduate :- not taken(c). }

\st (Here student is eliminated from the parameters and we are limited
to only one
required class, $c$.) Even though in this case the answers are correct,
unprincipled use of default negation leads to some potential
difficulties.
Suppose, for instance, that a student may graduate if
given a special permission. This can be naturally added as a rule

\st {\tt 
ready\_to\_graduate :- permitted. }

\st If the program is expanded by {\tt permitted} it becomes
inconsistent. This, of course, is unintended and contradicts our intuition. No such problem exists for the original representation.

\end{example}
The next example shows how the semantics deals with 
vicious circles. 
\begin{example}[Set atoms in the rule body]
Consider a program $P_4$

\st {\tt 
p(a) :- p $\subseteq$ \{X : q(X)\}. \\ 
q(a). }

\st in which definition of $p(a)$ depends on the existence of the set denoted by 
$\{X: p(X)\}$. In accordance with the vicious circle principle no answer 
set of this program can contain $p(a)$. There are only two candidates 
for answer sets of $P_4$: $S_1 = \{q(a)\}$ and $S_2 =
\{q(a),p(a)\}$. The set atom reduct of $P_4$ with respect to $S_1$
is 
\begin{verbatim} 
p(a) :- q(a). 
q(a). 
\end{verbatim}
while set atom reduct of $P_4$ with respect to $S_2$ is 
\begin{verbatim} 
p(a) :- p(a),q(a). 
q(a). 
\end{verbatim}
Clearly, neither $S_1$ nor $S_2$ is an answer set of $P_4$. As 
expected, the program is inconsistent. 
\end{example}
\subsubsection{Programs with Set Introduction Rules.}
A set introduction rule with head $p \subseteq S$ (where $p$ is a
predicate symbol and $S$ is a set name) defines set $p$
as an arbitrary subset of $S$; rule with head $p = S$ simply gives $S$
a different name; $S \subseteq p$ defines $p$ as an arbitrary superset
of $S$.
\begin{example}[Set introduction rule]\label{e20}
According to this intuitive reading the program $P_9$:

\st {\tt 
q(a). \\
p $\subseteq$ \{X:q(X)\}.}

\st has answer sets $A_1 = \{q(a)\}$ where the set $p$ is empty and 
 $A_2 = \{q(a),p(a)\}$ where $p = \{a\}$. 
\end{example}
The formal definition of answer sets of programs with set introduction
rules is given via a notion of 
\emph{set introduction reduct}. (The definition is similar to that 
presented in \cite{gel02}). 

\begin{definition}[Set Introduction Reduct]\label{reduct2}
The \emph{set introduction reduct} of a ground $\mathcal{A}log$ program $\Pi$ 
with respect to a set of 
ground regular literals $A$ is obtained from $\Pi$ by 
\begin{enumerate}
\item replacing every set introduction rule of $\Pi$ whose head is 
  not true in $A$ by 
$$\leftarrow body.$$
\item replacing every set introduction rule of $\Pi$ whose head $p
  \subseteq \{\bar{X}:q(\bar{X})\}$ (or $p = \{\bar{X}:q(\bar{X})\}$
  or  $\{\bar{X}:q(\bar{X})\} \subseteq p$) is 
  true in $A$ by 
$$p(\bar{t}) \leftarrow body$$
for each $p(\bar{t}) \in A$. 
\end{enumerate}
Set $A$ is an \emph{answer set} of $\Pi$ if it is an answer set of 
 the set introduction reduct of $\Pi$ with respect to $A$. 
\end{definition}
\begin{example}[Set introduction rule]\label{e25} 
Consider a program  $P_9$ from Example \ref{e20}. 
The reduct of this program with respect to $A_1 = \{q(a)\}$  is $\{q(a).\}$
and hence $A_1$ is an answer set of $P_9$. 
The reduct of $P_9$ with respect to  $A_2 = \{q(a),p(a)\}$ is 
$\{q(a). \ p(a).\}$ 
and hence $A_2$ is also an answer set of $P_9$. 
There are no other answer sets. 
\end{example}
The use of a set introduction rule  $p \subseteq S \leftarrow body$ is very
similar to that of choice rule $\{p(\bar{X}) : q(\bar{X})\} \leftarrow body$ of \cite{nss02} implemented in Clingo and
other similar systems. 
In fact, if $p$ from the set introduction rule 
 does not occur in the head of any other rule of the program,
the two rules have the same meaning.
However if this condition does not hold the meaning is different. An 
$\mathcal{A}log$ program consisting of rules $p \subseteq \{X:q_1(X)\}$
and $p \subseteq \{X:q_2(X)\}$ defines an arbitrary set $p$ from the 
  intersection of $q_1$ and $q_2$. With choice rules it is not the case.
We prefer the set introduction rule because of its more intuitive reading 
(after all everyone is familiar with
the statement ``$p$ is an arbitrary subset of $q$'') and relative
simplicity of the definition of its formal semantics as compared with
that of the choice rule. 

\st
Our last example shows how subset introduction rule with equality can 
be used to represent synonyms:
\begin{example}[Synonyms]\label{e26} 
Suppose we have a set of cars represented by atoms formed by a 
predicate symbol $car$, e.g., $\{car(a).\ car(b).\}$
The following rule 
\begin{verbatim}
carro = {X:car(X)} :- spanish. 
\end{verbatim}
allows to introduce a new name of this set for Spanish speaking 
people. 
Clearly, $car$ and $carro$ are synonyms. Hence, program $P_9 \cup
\{spanish.\}$ has one answer set:
$\{spanish, car(a), car(b), carro(a),carro(b)\}$. 
\end{example}
\section{Alternative Formalization of VCP -- Language $\mathcal{S}log^+$}
In this section we introduce alternative interpretation of VCP
(referred to as \emph{weak VCP}) and incorporate it in the
semantics of a new logic programming language with
set, called $\mathcal{S}log^+$. The syntax of $\mathcal{S}log^+$ coincides with that of
$\mathcal{A}log$. Its informal semantics is based on weak VCP. 
By $C(T)$ we denote a set atom containing an occurrence of set term $T$.
The {\em instantiation} of $C(\{X:p(X)\})$ in a set $A$ of regular literals
obtained from $C(\{X:p(X)\})$ by replacing
$\{X:p(X)\}$ by $\{t:p(t) \in A\}$. The weak VCP is: 
{\em belief in p(t) (i.e. inclusion of p(t) in an answer set $A$) must be established
without reference to the instantiation of a set atom $C$ in $A$ unless the truth of this instantiation can be demonstrated without reference to $p(t)$.}

\begin{example}\label{e31}
{\rm To better understand the weak VCP, let us consider program}

\begin{verbatim}  
p(0) :- C.
:- not p(0).
\end{verbatim}
 
\noindent First we assume $C$ be $card\{X:p(X)\} > 0$. There is only one candidate answer set $A = \{p(0)\}$ for this program.
 Belief in $p(0)$ (i.e. its membership in answer set $A$) can only be established
by checking if instantiation $card\{t : p(t) \in A\} > 0$ of $C$ in $A$ holds.
This is prohibited by weak VCP unless the truth of this instantiation can be demonstrated
without reference to $p(0)$. But this cannot be so demonstrated because $card\{t : p(t) \in A\} > 0$ holds 
only when $p(0)$ is in $A$. Hence, $A$ is not an answer set.  
Now let $C$ be $card\{X:p(X)\} \ge 0$.
This time the truth of instantiation $card\{t : p(t) \in A\} \ge 0$ of $C$ can be demonstrated without reference to $p(0)$ -- the instantiation would be true even if $A$ were empty. Hence $p(0)$ must be believed and thus the program
has one answer set, $\{p(0)\}$.
\end{example}

\hide{
\begin{example}\label{e31}
{\rm To better understand the weak VCP let us consider programs in
Example \ref{e1} from the standpoint of $\mathcal{S}log^+$.
Inconsistency of $P_0$ is not caused by self-reference and the program remains
inconsistent under the new semantics.
As before, there are two candidate answer sets of $P_1$:
$A_1=\emptyset$ and $A_2=\{p(0),p(1)\}$. 
$A_1$ is ruled out by the satisfiability requirement. $A_2$ was shown
to violate the original version of VCP but one can see that it
does not satisfy weak VCP either.
Beliefs in $p(0)$ and $p(1)$ depend on the existence of the set $S =
\{t:p(t) \in A_2\} = \{0,1\}$ satisfying condition $card(S) \not= 1$.
The truth of this condition cannot be however established without
reference to these beliefs.
So $P_1$ is still inconsistent. 
For $P_2$ the situation changes. Consider a candidate answer set $A=\{p(1)\}$. 
In this case $S = \{t:p(t) \in A\} = \{1\}$. Truth of
condition $card(S) \geq 0$ does not depend on $p(1)$.
Hence the set can be formed without violating
the weak VCP and $\{p(1)\}$ is the only answer set of $P_2$.
}
\end{example}
}

To make weak VCP based semantics precise we need the following notation and definitions:
By
$\bar{W}^n, \bar{V}^n$ we denote n-ary vectors of sets of ground
regular literals and by $W_i$, $V_i$
their $i$-th coordinates.
$\bar{W}^n \leq \bar{V}^n$ if for every $i$, $W_i \subseteq V_i$.
$\bar{W}^n < \bar{V}^n$ if $\bar{W}^n \leq \bar{V}^n$ and $\bar{W}^n \not= \bar{V}^n$.
A set atom $C(\{X : p_1(X)\},\dots,\{X : p_1(X)\})$
is \emph{satisfied} by $\bar{W}^n$ if
$C(\{t : p_1(t) \in W_1\}, \dots, \{t : p_n(t) \in W_n\})$
is true.

\begin{definition}[Minimal Support]\label{d1}
Let $A$ be a set of ground regular literals of $\Pi$, and $C$ be a set atom with
$n$ parameters. $\bar{W}^n$ 
is a \emph{minimal support}
for $C$ in $A$ if
\begin{itemize}
\item 
For ever $1 \leq i \leq n$, $W_i \subseteq A$.
\item Every $\bar{V}^n$ 
such that for every $1 \leq i \leq n$, $W_i \subseteq V_i \subseteq A$ satisfies $C$. 

\item No $\bar{U}^n < \bar{W}^n$ satisfies the first two conditions. 

\end{itemize}

\end{definition}
Intuitively, the weak VCP says that set atom $C$ can be safely used to support the 
reasoner's beliefs iff the existence of a minimal support of $C$ can be established without
reference to those beliefs.
Precise definition of answer sets of $\mathcal{S}log^+$ is obtained
by replacing definition \ref{reduct1} of set reduct of
$\mathcal{A}log$ by definition \ref{vcp2} below and combining it with
definition \ref{reduct2}.
\begin{definition}[Set-reduct of $\mathcal{S}log^+$] \label{vcp2}
A \emph{set reduct} of $\mathcal{S}log^+$ program $\Pi$ with respect to a set $A$ of 
ground regular literals is obtained from $\Pi$ by 
\begin{enumerate}
\item Removing rules containing set atoms
  which are \emph{false} or \emph{undefined} in $A$. 
\item 
Replacing every remaining set atom $C$ in the body of the rule by
the union of coordinates of one of its minimal supports.
\end{enumerate}
Clearly such a reduct is a regular ASP program without sets. 
$A$ is an \emph{answer set} of a $\mathcal{S}log^+$ program $\Pi$ 
if $A$ is an answer set of a weak 
set reduct
of $\Pi$ with respect to $A$. 
\end{definition}



\begin{example}\label{e4}
{\rm 
Consider now an $\mathcal{S}log^+$ program $P_3$
\begin{verbatim}
p(3) :- card{X : p(X)} >= 2. 
p(2) :- card{X : p(X)} >= 2. 
p(1). 
\end{verbatim}
It has two candidate answer sets: $A_1=\{p(1)\}$ and
$A_2=\{p(1),p(2),p(3)\}$. In $A_1$ the corresponding condition is not satisfied
and, hence, the weak set reduct of the program with respect to $A_1$ is
$p(1).$ Consequently, $A_1$ is an answer set of $P_3$. In $A_2$ 
the condition has three minimal supports: $M_1 = \{p(1),p(2)\}$, $M_2
= \{p(1),p(3)\}$, and $M_3 = \{p(2),p(3)\}$. Hence, the program has
nine weak set reducts of
$P_3$ with respect to $A_2$. Each reduct is of the form
\begin{verbatim}
p(3) :- Mi.
p(2) :- Mj.
p(1).
\end{verbatim}
where $M_i$ and $M_j$ are minimal supports of the condition.
Clearly, the first two rules of such a reduct are useless and hence
$A_2$ is not an answer set of this reduct. Consequently $A_2$ is not
an answer set of $P_3$.
}
\end{example}


\hide{For program $P_0$, consider candidate answer set $A = \{p(0),p(1)\}$. $A$ is a minimal support for {\tt card\{X:p(X)\} != 1} in $A$.
The set-redutt of $P_0$ wrt $A$ is  \{{\tt p(1) :- p(0), p(1)}\} for which 
$A$ is not the answer set. So, $A$ is not an answer set of $P_0$. Similarly, there is no answer set for $P_1$. For $P_2$, consider the candidate answer set $A = \{p(1)\}$. $\{\}$ is a minimal support for {\tt card\{X:p(X)\} $\ge$ 0} in $A$. 
The set-reduct of $P_2$ wrt $A$ is 
\{\noindent {\tt p(1)}\}.
Hence, $A$ is an answer set of $P_2$. 
}

\st The following two results help to better understand the semantics of 
$\mathcal{S}log^+$.
\begin{theorem}\label{th1}
If a set $A$ is an $\mathcal{A}log$ answer set of $\Pi$ then $A$ is an 
$\mathcal{S}log^+$ answer set of $\Pi$. 
\end{theorem} 
As an $\mathcal{S}log^+$  program, $P_2$ has an answer set of $\{p(1)\}$, but it has no answer set as an $\mathcal{A}log$ program.
The following result shows that there
are many such programs and justifies our name for the new language.

\begin{theorem}\label{th2}
Let $\Pi$ be a program which, syntactically, belongs to both $\mathcal{S}log$ and $\mathcal{S}log^+$. A set $A$ is an $\mathcal{S}log$ answer set of $\Pi$ iff it is an 
$\mathcal{S}log^+$ answer set of $\Pi$. 
\end{theorem} 


As shown in \cite{SonPT07} $\mathcal{S}log$ has sufficient expressive power to formalize
complex forms of recursion, including that used in the Company Control
Problem \cite{FaberPL11}.
Theorem \ref{th2} guarantees that the
same representations will work in $\mathcal{S}log^+$.
Of course, in many respects $\mathcal{S}log^+$ substantially increases
the expressive power of $\mathcal{S}log$. Most importantly it expands the 
 $\mathcal{S}log$ semantics to programs with epistemic disjunction --
 something which does not seem to be easy to do using the original
 definition of $\mathcal{S}log$ answer sets. Of course, new set
 constructs and rules with infinite number of literals are available
 in $\mathcal{S}log^+$ but not in $\mathcal{S}log$. On another hand,
 $\mathcal{S}log$ allows multisets --  a feature we were not trying to include
 in our language.
 The usefulness of multisets and the analysis of its cost in
 terms of growing complexity of the language due to its introduction
 is still under investigation.

\medskip
Unfortunately, the additional power of $\mathcal{S}log^+$ as compared with 
 $\mathcal{A}log$ comes at a price. Part of it is a comparative
 complexity of the definition of $\mathcal{S}log^+$ set reduct. But, more importantly,
 the formalization of the weak VCP does not eliminate
 all the known paradoxes of reasoning with sets.
Consider, for instance the following example:
\begin{example}\label{e5}
{\rm 
Recall program $P_2$:
\begin{verbatim}
p(1) :- card{X:p(X)} >= 0.
\end{verbatim}
from Example \ref{e1} and assume, for simplicity, that parameters of
$p$ are restricted to $\{0,1\}$.
Viewed as a program of $\mathcal{A}log$, $P_2$ is inconsistent. In
$\mathcal{S}log^+$ (and hence in $\mathcal{S}log$ and
$\mathcal{F}log$ (the language defined in \cite{FaberPL11})) it has an answer set $\{p(1)\}$. 
The latter languages therefore admit existence of set $\{X:p(X)\}$.
Now let us look at program $P_5$:
\begin{verbatim}
p(1) :- card{X : p(X)} = Y, Y >=0.
\end{verbatim}
and its grounding $P_6$:
\begin{verbatim}
p(1) :- card{X:p(X)} = 1, 1>=0. 
p(1) :- card{X:p(X)} = 0, 0>=0. 
\end{verbatim}
They seem to express the same thought as $P_2$, and it is natural to
expect all these programs to be equivalent. It is indeed true in
$\mathcal{A}log$ -- none of the programs is consistent. 
According to the semantics of $\mathcal{S}log^+$ (and $\mathcal{S}log$ and $\mathcal{F}log$), however, $P_5$ and
$P_6$ are inconsistent. To see that notice that there are two
candidate answer sets for $P_6$: $A_1=\emptyset$ and $A_2 = \{p(1)\}$.
The minimal support of $card\{X:p(X)\}
= 0$ in $A_1$ is $\emptyset$ and hence the only weak set
reduct of $P_6$ with respect to $A_1$ is \{{\tt p(1) :- 0>=0}\}.
$A_1$ is not an answer set of $P_6$. The minimal support of $card\{X:p(X)\}=1$
in $A_2$ is $\{p(1)\}$.
The only weak set reduct is \{
{\tt p(1) :- p(1),1>=0}
\}.
$A_2$ is not an answer set of $P_6$ either. It could be that
this paradoxical behavior will be in the future explained from some basic
principles but currently authors are not aware of such an explanation.

}
\end{example}

\section{Properties of VCP Based Extensions of ASP}

In this section we give some basic properties of $\mathcal{A}log$ and $\mathcal{S}log^+$ programs. 
Propositions \ref{p1aa} and \ref{p2aa} ensure that, as in regular ASP,
answer sets of $\mathcal{A}log$ program are formed using the program 
rules together with the rationality principle. 
Proposition \ref{split} is the $\mathcal{A}log$/$\mathcal{S}log^+$ version of the 
Splitting Set Theorem -- basic 
technical tool used in theoretical investigations of ASP and its 
extensions \cite{GelfondP92,lt94,Turner96}. 

\begin{proposition}[Rule Satisfaction and Supportedness]\label{p1aa} Let 
  $A$ be an $\mathcal{A}log$ or $\mathcal{S}log^+$ answer set of a ground program $\Pi$. 
Then 
\begin{itemize}
  \item $A$ satisfies every rule $r$ of $\Pi$. 
  \item If $p(\bar{t}) \in A$ then there is a rule $r$ from $\Pi$ such
    that the body of $r$ is satisfied by $A$ and
\begin{itemize}
\item 
    $p(\bar{t})$ is the only atom 
    in the head of $r$ which is true in $A$ or
\item  the head of $r$ 
is of the form $p \odot \{\bar{X} : q(\bar{X})\}$ and $q(\bar{t}) \in A$. 
    (It is often said that 
    rule $r$ supports atom $p$.) 
\end{itemize}
\end{itemize}
\end{proposition}

\noindent By the intuitive and formal meaning of set introduction
rules, the anti-chain property no longer holds. 
However, the anti-chain property still holds for programs without set
atoms in the heads of their rules.

\begin{proposition}[Anti-chain Property]\label{p2aa}
If $\Pi$ is a program without set
atoms in the heads of its rules then there are no $\mathcal{A}log$ answer sets $A_1$,
$A_2$ of $\Pi$ such that $A_1 \subset A_2$. Similarly for its
$\mathcal{S}log^+$ answer sets.

\end{proposition}
Before formulating the next result we need some terminology. 

\begin{definition}[Occurrences of Regular Literals in Aggregate Atoms]\label{occur}
We say that a ground literal $l$ \emph{occurs in a set atom} $C$ if there is a set name $\{X:cond(X)\}$ occurring in $C$ and $l$ is a ground instance of some literal in 
$cond$. If $B$ is a set of ground literals possibly preceded by 
default negation $not$ then $l$ occurs in $B$ if $l \in B$, or 
$not\ l \in B$, or $l$ occurs in some set atom from $B$. 
\end{definition}
\begin{definition}[Splitting Set]\label{split-set}
Let $\Pi$ be a program with signature $\Sigma$. 
A set $S$ of ground regular literals of $\Sigma$ is called a 
\emph{splitting set} of $\Pi$ if, for every rule $r$ of $\Pi$, 
if $l$ occurs in the head of $r$ then every literal occurring in the 
body of $r$ belongs to $S$. 
The set of rules of $\Pi$ constructed from literals of $S$ is called 
\emph{the bottom} of $\Pi$ relative to $S$; the remaining rules 
are referred to as \emph{the top} of $\Pi$ relative to $S$.  
\end{definition}
Note that the definition implies that no literal occurring in the 
bottom of $\Pi$ relative to $S$ can occur in the heads of rules from 
the top of $\Pi$ relative to $S$.

\begin{proposition}[Splitting Set Theorem]\label{split}
Let $\Pi$ be a ground program, 
$S$ be its splitting set, and $\Pi_1$ and $\Pi_2$ be the bottom and 
the top of $\Pi$ relative to $S$ respectively. Then 
a set $A$ is an answer set of $\Pi$
iff $A \cap S$  is an answer set of $\Pi_1$ and $A$ is an answer set 
of $(A \cap S) \cup \Pi_2$. 
\end{proposition}
Note that this formulation differs from the original one in two 
respects. First, rules of the program can be infinite. Second,
the definition of occurrence of a regular literal
in a rule changes to accommodate the presence of set atoms. 


\section{Related Work}
There are multiple approaches to introducing aggregates in logic 
programming languages under the answer sets semantics 
\cite{KempS91,gel02,nss02,Marek04,Marek2004set,Pelov04,PelovDB04,PelovT04,Ferraris05,FerrarisL05,pdb07,SonP07,LeeLP08,ShenYY09,LiuPST10,FaberPL11,PontelliST11,liu2011strong,WangLZY12,HarrisonLY14,shen2014flp,GelfondZ14,gebser2015abstract,alviano2015complexity}. 
In addition to this work our paper was significantly influenced by the 
original work on VCP in set theory and principles of language design 
advocated by Dijkstra, Hoare, Wirth and others. Harrison et al's  work \cite{HarrisonLY14}
explaining the semantics of some constructs of gringo in terms of 
infinitary formulas of Truszczynski \cite{truszczynski2012connecting} led to their inclusion in  $\mathcal{A}log$ and $\mathcal{S}log^+$. 
The notion of set reduct of $\mathcal{A}log$ was influenced by the reduct introduced for defining the semantics of Epistemic Specification
in \cite{Gelfond2011new}. Recent work by Alviano and Faber
\cite{alviano2015stable} helped us to realize the close 
relationship between  $\mathcal{A}log$ and $\mathcal{S}log$ and Argumentation theory \cite{dung1995acceptability,brewka2013abstract,strass2013approximating} which certainly deserves further investigation, as well as 
provided us with additional knowledge about  $\mathcal{A}log$. More information about $\mathcal{S}log$ and $\mathcal{S}log^+$ can be found in Section 3. 
\hide{
We start with a short discussion of our semantics for ASP with
infinite rules. The idea of expending the syntax of ASP to allow
infinitary formulas is not new. Stable model semantics for
such formulas was first
introduced in \cite{Truszczynski12}. It is based on the notion of stable models
for finite propositional formulas from \cite{Ferraris05,Ferraris11}. 
Proposition 28 from \cite{fl05} shows that, for finite ASP rules, the
latter definition coincides with the original definition of answer
sets. The proof of this proposition can be easily adopted to the
infinite case to show that a semantics of ASP program with infinite
rules coincides with that from \cite{Truszczynski12}.
}
Shen et al. \cite{ShenYY09} and Liu et al. \cite{liu2011strong}
propose equivalent semantics for disjunctive constraint programs (i.e., programs with rules whose bodies are built from constraint atoms and whose heads are epistemic disjunctions of such atoms). This generalizes 
the standard ASP semantics for disjunctive programs. We conjecture that when we adapt our definition of $\mathcal{S}log^+$ semantics to disjunctive constraint programs, it will coincide with that of \cite{ShenYY09,liu2011strong}. 
However, our definition seems to be simpler and is based on clear, VCP related intuition. 

\section{Conclusion}
The paper belongs to the series of works aimed at the development of an
answer set based knowledge representation language. Even though we
want to have a language suitable for serious applications our main
emphasis is on teaching. This puts additional premium on clarity
and simplicity of the language design. In particular we believe that
the constructs of the language should have a simple syntax and 
 a clear intuitive semantics based on understandable informal principles. 
In our earlier paper \cite{GelfondZ14} we concentrated on a language
$\mathcal{A}log$ expanding standard Answer Set Prolog by
aggregates. We argued that the syntax of the language is simpler than that
of the most popular aggregate language $\mathcal{F}log$ implemented in
Clingo and other similar systems. In particular, $\mathcal{A}log$'s
notion of grounding \emph{allows to define the intuitive (and formal) meaning of a set name
independently from its occurrence in a rule}. As the result, set name
$\{X : p(X)\}$ can be always equivalently replaced by $\{Y : p(Y)\}$.
In $\mathcal{F}log$, it is not the case. A semantics of aggregates in
$\mathcal{A}log$ was based on a particularly simple and restrictive
formalization of VCP. In this paper we:
\begin{itemize}
\item Expanded syntax and semantics of the original $\mathcal{A}log$
  by allowing
\begin{itemize}
\item 
  rules with an infinite number of literals -- a feature of
  theoretical interest also useful
  for defining aggregates on infinite sets;
\item  subset relation between sets in the bodies of rules concisely
  expressing a specific form of universal quantification;
\item  set introduction -- a feature with functionality somewhat
  similar to that of the choice rule of clingo but with different
  intuitive semantics.  
\end{itemize}
Our additional set constructs are aimed at showing that our
original languages can be expanded in a natural and technically simple
ways.
Other constructs such as set operations
and rules with variables ranging over sets (in the style of
\cite{DovierPR03}), etc. are not discussed. Partly this is due to
space limitations -- we do not want to introduce any new constructs
without convincing examples of their use. The future will show if such
extensions are justified.
 
\item Introduced a new KR language, $\mathcal{S}log^+$, with the same
  syntax as $\mathcal{A}log$ but different semantics for the set
  related constructs. The new language is less restrictive and allows
  formation of substantially larger collection of sets. Its semantics
  is based on the alternative, weaker formalization of VCP. 
\item Proved that (with the exception of multisets)
  $\mathcal{S}log^+$ is an extension of a well known aggregate language $\mathcal{S}log$. 
The semantics of
the new language is based on the intuitive idea quite different from
that of $\mathcal{S}log$ and the definition of its semantics is
simpler. We point out some paradoxes of $\mathcal{S}log^+$ (and $\mathcal{F}log$)
which prevent us from advocating them as standard ASP language with aggregates.

\item Proved a number of basic properties of programs of
  $\mathcal{A}log$ and $\mathcal{S}log^+$.
\end{itemize}

\bibliographystyle{splncs03}
\bibliography{biblio}
\end{document}